\documentclass[conference]{IEEEtran}
\usepackage{cite}
\usepackage{amsmath,amssymb,amsfonts}
\usepackage{algorithmic}
\usepackage{graphicx}
\usepackage{textcomp}
\usepackage[table,xcdraw]{xcolor}
\usepackage{tabularx}
\usepackage{float}
\usepackage{multirow}
\usepackage{hhline}
\usepackage{siunitx}
\usepackage{textcomp}
\usepackage{enumitem}

\newcolumntype{C}[1]{>{\centering\arraybackslash\hspace{0pt}}p{#1}}
\usepackage{color}

\usepackage[utf8]{inputenc}
\usepackage[htt]{hyphenat}

\usepackage{comment}
\usepackage[export]{adjustbox}

\usepackage{listings}
\usepackage{csquotes}

\usepackage[hidelinks,bookmarks=false]{hyperref}

\begin{document}

\title{The Impact of Data Preparation on the \\ Fairness of Software Systems}

\author{
	\IEEEauthorblockN{Inês Valentim, Nuno Lourenço, Nuno Antunes}
	\IEEEauthorblockA{\textit{CISUC, Department of Informatics Engineering}\\
	\textit{University of Coimbra}\\
	Coimbra, Portugal \\
	\href{valentim@dei.uc.pt}{valentim@dei.uc.pt}, 
	\href{naml@dei.uc.pt}{naml@dei.uc.pt}, 
	\href{nmsa@dei.uc.pt}{nmsa@dei.uc.pt}}
}

% \IEEEspecialpapernotice{\color{red}{(PER: Practical Experience Report)}}

\maketitle

% to be removed for camera ready
% \thispagestyle{plain}
% \pagestyle{plain}

\begin{abstract}
Machine learning models are widely adopted in scenarios that directly affect people. 
The development of software systems based on these models raises societal and legal concerns, as their decisions may lead to the unfair treatment of individuals based on attributes like race or gender.
% Data preparation is key in any machine learning pipeline, including instance selection and data transformations, but its effect on fairness is yet to be studied in detail.
Data preparation is key in any machine learning pipeline, but its effect on fairness is yet to be studied in detail.
In this paper, we evaluate how the fairness and effectiveness of the learned models are affected by the removal of the sensitive attribute, the encoding of the categorical attributes, and instance selection methods (including cross-validators and random undersampling).
% We used the Adult Income and the German Credit Data datasets, which are widely known and have fairness concerns.
We used the Adult Income and the German Credit Data datasets, which are widely studied and known to have fairness concerns.
% Each data preparation technique was applied individually, and we measured the difference on the effectiveness metrics and on fairness metrics such as statistical parity difference, disparate impact, and normalised prejudice index.
We applied each data preparation technique individually to analyse the difference in predictive performance and fairness, using statistical parity difference, disparate impact, and the normalised prejudice index.
The results show that fairness is affected by transformations made to the training data, particularly in imbalanced datasets. 
Removing the sensitive attribute is insufficient to eliminate all the unfairness in the predictions, as expected, but it is key to achieve fairer models. 
% Applying a sampling method due to data imbalance also impacts fairness, with the standard random undersampling with respect to the true labels being sometimes more prejudicial than performing no random undersampling.
Additionally, the standard random undersampling with respect to the true labels is sometimes more prejudicial than performing no random undersampling.
\end{abstract}

\begin{IEEEkeywords}
Fairness, 
%Discrimination, 
Data Preparation,
Machine Learning.
\end{IEEEkeywords}

\vspace{-10pt}

\section{Introduction}
\label{sec:intro}

Software systems based on machine learning (ML) are being used at an increasingly higher rate and on a multitude of scenarios that have a significant impact on people's lives. 
Their ubiquity raises several legal and societal concerns, as decisions based on the output of ML models may introduce or perpetuate historical bias against some individuals, based on their intrinsic characteristics, such as race, gender or age. 
The use of automated decision-making systems is often appealing due to the gains associated with it, and might even be perceived as a step towards the eradication of personal bias from the process.
Nevertheless, many are the risks associated with a careless adoption of decisions supported by these systems. 

In this context, fairness emerges as a key property in terms of the reliability and trustworthiness of software systems based on ML. 
These receive nowadays increased attention from regulatory institutions, with the recently approved European Union General Data Protection Regulation (GDPR) demanding organisations to handle personal data in a privacy-preserving, fair and transparent manner~\cite{EU_GDPR}.

Techniques to assess fairness and build models capable of providing fairer predictions are of great help to organisations which intend to be GDPR compliant, but may lack the resources or knowledge~\cite{dsml_2018}. 
These organisations must be aware of the potential biases in their models at the design, implementation and deployment phases, and should make regular fairness evaluations of their systems~\cite{ACM:transparencystatement}. 
% Moreover, the assessment approaches may be used to perform audits of non-compliant organisations, therefore providing valuable insight on how they are violating these fairness principles~\cite{dsml_2018}.
Moreover, the assessment approaches may be used to audit non-compliant organisations, therefore providing valuable insight on violations of these fairness principles~\cite{dsml_2018}.
Individuals who rely on these organisations also benefit from the deployment of fairness-aware models and the adoption of such practices, since they provide an extra assurance that their data is not being used in ways that may negatively impact their daily lives.

% \toiv{este paragrafo abaixo parece nao adicionar muito. Deve ser substituirdo pelo que falo a seguir}
% As real people are already making decisions with the support of ML, sometimes with the system's design and implementation not taking these concerns into consideration, it is crucial to understand if widely adopted ML techniques jeopardise its fairness. Therefore, we focus on the \textbf{assessment of the impact of data preparation and pre-processing procedures applied to a dataset on the fairness of systems based on machine learning}.
% \toiv{Falta aqui um paragrafo em que se diz o que foi feito}
% Many learning algorithms (or classifiers) can be applied in a supervised classification setting. For this work, we focused on \textbf{tree-based methods}, which encompass Decision Trees and Random Forests, in part due to the models resulting from their application being more easily interpretable. For this reason, they can be classified as grey-box methods.
In this paper, we \textbf{assess the impact of widely adopted data preparation procedures on the fairness of systems based on ML}. More precisely, we consider the removal of the sensitive attribute, the encoding of the categorical attributes, and instance selection methods, like cross-validation and random undersampling.
Despite not being in the main scope of this work, we also consider the influence of the learning algorithm on fairness.
From the many algorithms suitable for a supervised classification setting, we first focus on tree-based methods, like Decision Trees and Random Forests, partly due to the easier interpretability of the resulting models.

The obtained results show the importance of adopting the standard legal practices to mitigate discrimination, namely the removal of the sensitive attribute prior to training. 
However, we have also found that this procedure might not always lead to the expected behaviour, with the models' predictions sometimes being more unfair than when the model has access to the sensitive attribute. 
We also report the drawbacks of using more complex learning algorithms, with Random Forests making more discriminatory predictions than Decision Trees. 
Furthermore, we emphasise that caution must be taken when dealing with datasets which show an imbalance with respect to both the true labels and the sensitive attribute. 
Standard sampling methods, such as random undersampling with respect to the true labels, may have undesired effects on fairness.

The remainder of this paper is organised as follows. 
Section~\ref{sec:background-related-work} provides an overview of the key concepts of machine learning and fairness, and reviews related work on fairness of software systems based on machine learning models. 
Section~\ref{sec:methodology} presents the research questions and details the experimental methodology.
The obtained results are presented in Section~\ref{sec:results} and our findings are discussed in Section~\ref{sec:discussion}. 
Section~\ref{sec:conclusion} concludes the paper.

\section{Background and Related Work}
\label{sec:background-related-work}

%The ubiquity of \textit{artificial intelligence} led to the usage of machine learning models to support real-world decisions. 
%In many of these contexts, the decisions directly or indirectly impact the life of individuals, therefore opening doors to several legal and societal concerns to arise, some of which related to discrimination and fairness. 
In this section we overview machine learning concepts key to our work, after which we present the core concepts of fairness and related work on the topic, namely development of fairness-aware algorithms and fairness metrics.

\subsection{Machine Learning}
\label{ssec:machine-learning}
% The aim of \textbf{machine learning (ML)} is to enable software systems to learn from data, i.e. to make them modify and adapt their actions so that these actions become more accurate (closer to the desired outcome)~\cite{Marsland}. Our focus is on \textbf{supervised learning}: we have access to training examples which include not only the features (or attributes), but also the outcome variable (targets or true labels), used to guide the learning process~\cite{Bishop2006,ESL:HastieTF09}. Moreover, we will be dealing with \textbf{classification problems}: we want to assign each instance to one of the possible classes, and so the outcome variable is discrete. For instance, we might want to determine whether someone has a high or low risk of committing another crime.
Machine learning (ML) aims at enabling software systems to learn from data, by modifying and adapting their actions towards the desired outcome~\cite{Marsland}.
Our focus is on \textbf{supervised classification problems}: we have access to the features (or attributes) of the training instances and to the \textit{discrete outcome variable} (or true labels), which guides the learning process~\cite{Bishop2006,ESL:HastieTF09}. The goal is to assign one of the possible classes to each instance. For example, we may want to determine whether someone has a high or a low risk of recidivism.

\begin{figure}[htbp]
	\begin{center}
    \includegraphics[width=1.0\linewidth]{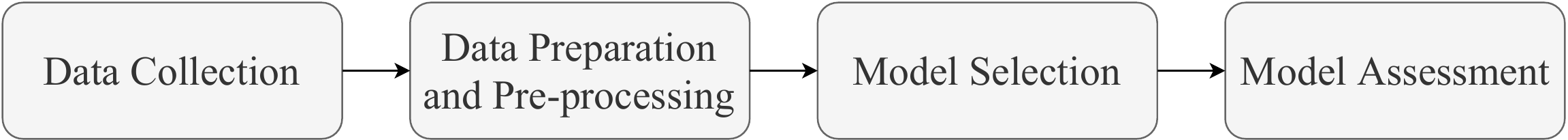}
    \vspace{-10pt}
    \caption{Typical machine learning pipeline.}
	\label{fig:ml-pipeline}
	\end{center}
	\vspace{-10pt}
\end{figure}

% A system which relies on ML usually follows a pipeline as shown in Fig.~\ref{fig:ml-pipeline}: after the data is collected, it goes through a set of data preparation and pre-processing steps, followed by the model selection and assessment phases.
A system based on ML usually follows a pipeline as shown in Fig.~\ref{fig:ml-pipeline}. The \textbf{data collection} phase includes gathering representative data for the problem at hand, as well as labelling the training examples when in the presence of a supervised learning task. The \textbf{data preparation and pre-processing} steps may include handling missing data, encoding categorical features, discretisation, feature normalisation, feature selection and feature reduction techniques.
% Not only is the application of these techniques of pivotal importance for some models to deliver the expected results, but it also helps dealing with overfitting.
It is crucial to apply these techniques for models to deliver the expected results, while helping to deal with overfitting.
\textbf{Model selection} deals with the process of selecting the most appropriate model for the problem we are trying to solve, taking the complexity and flexibility of the models into account~\cite{ISLR}. \textbf{Model assessment} deals with evaluating the performance of the chosen model by estimating its generalisation error on new unseen data~\cite{ESL:HastieTF09,ISLR}.

% \toiv{Talvez seja de limitar o scope to trees tambem na intro, nao? Talvez naquele paragrafo novo}
% Many learning algorithms (or classifiers) can be applied in a supervised classification setting. For this work, we focused on \textbf{tree-based methods}, which encompass Decision Trees and Random Forests, in part due to the models resulting from their application being more easily interpretable. For this reason, they can be classified as grey-box methods. \textbf{Decision Trees} try to learn simple decision rules from the available features in the training data~\cite{scikit-learn}. A classification tree is built by following a recursive binary splitting process guided by a criterion which evaluates the quality of the splits~\cite{ISLR}. Common choices for this criterion include the classification error rate, the Gini index and cross-entropy~\cite{ISLR}. Tree pruning can be used to avoid overfitting. Some well-known decision tree algorithms include ID3 and C4.5.
\textbf{Decision Trees} try to learn simple decision rules from the features of the training data~\cite{scikit-learn}. A classification tree is built by following a recursive binary splitting process guided by the evaluation of the splits' quality using a criterion like the classification error rate, the Gini index, or cross-entropy~\cite{ISLR}. Tree pruning can be used to avoid overfitting. Some well-known decision tree algorithms include ID3 and C4.5.
\textbf{Random Forests} are collections of decision trees where the final prediction is given by a majority vote over the predictions of all the trees in the ensemble~\cite{Shalev-Shwartz2014}. To reduce the correlation between the trees, the candidates for splitting are randomly selected from the full set of input features before each split~\cite{ESL:HastieTF09}. This randomisation process also aims at reducing variance~\cite{ISLR}.

% To choose a model for the problem we are trying to solve, we have to assess its generalisation performance, which is related to the model's capability of making accurate predictions given new unseen instances~\cite{ESL:HastieTF09}. A \textbf{training set} should be used to fit the models; a \textbf{validation set} should be used to estimate the performance of different models so as to choose the best one; and a \textbf{test set} should be used to assess the generalisation error of the chosen model~\cite{ESL:HastieTF09}. By eliminating the need to set aside a validation set, \textbf{cross-validation} is a commonly adopted procedure when there is insufficient data to partition the dataset into three sets.

% \toiv{este paragrafo não tem mais do que 2 frases essenciais. é preciso avaliar capacidade generativa. Ha cenas tais como cross validation.}
% To choose a model for the problem we are trying to solve, we have to assess its generalisation performance (\na{a.k.a. generativate capacity}), which is related to its capability of making accurate predictions given new unseen instances~\cite{ESL:HastieTF09}. 
% Ideally, a dataset should be partitioned into a training set (used to fit the models), a validation set (used to estimate the performance of different models so as to choose the best one), and a test set (used to estimate the generalisation error of the chosen model)~\cite{ESL:HastieTF09}. 
% By eliminating the need to set aside a validation set, \textbf{cross-validation} is commonly adopted when there is insufficient data to make this partition.
To choose a model we need to assess its generalisation performance (capability of making accurate predictions given new unseen instances)~\cite{ESL:HastieTF09}.
\textbf{Cross-validation} is a suitable approach when there is insufficient data to make a partition into training, validation, and test sets.
Furthermore, the choice of a performance metric is dependent on the problem and the characteristics of the available data. 
A \textit{confusion matrix}, as shown in Table~\ref{tab:confusion-matrix}, summarises the results of a binary classification problem, with four possible classification results.
Several performance metrics, whose definitions can be found in~\cite{Marsland}, can be derived from this matrix.

% \toiv{Esta parte para baixo precisa de encolher MUITO}

% The choice of a \textbf{performance metric} is always dependent on the problem and the characteristics of the available data. In what follows, we will assume binary classification problems. A \textbf{confusion matrix}, as represented in Table~\ref{tab:confusion-matrix}, summarises the results of a classification problem in a tabular format, where rows correspond to the true class and columns correspond to the predicted class. Each instance may fall in one of four possible classification results: a True Positive (TP) and a False Negative (FN) are a positive instance correctly and incorrectly classified, respectively; while a False Positive (FP) and a True Negative (TN) are a negative instance incorrectly and correctly classified, respectively. Several performance metrics, whose definitions can be found in~\cite{Marsland}, can be derived from the confusion matrix.
% A true positive (TP) and a false negative (FN) are a positive instance correctly and incorrectly classified, respectively; while a false positive (FP) and a true negative (TN) are a negative instance incorrectly and correctly classified, respectively. 

\vspace{-6pt}

\begin{table}[htbp]
    \caption{Confusion matrix for a binary classification problem.}
    \centering
    \renewcommand{\arraystretch}{1.15}
    \begin{tabular}{c|c|c|c|}
    \hhline{~|~|-|-|}
    % \cline{3-4}
    \multicolumn{2}{c|}{}                                                                 & \multicolumn{2}{c|}{\cellcolor[HTML]{C0C0C0}Predicted Class} \\ \cline{3-4} 
    \multicolumn{2}{c|}{\multirow{-2}{*}{}}                                               & Positive                      & Negative                     \\ \hline
    \multicolumn{1}{|c|}{\cellcolor[HTML]{C0C0C0}}                             & Positive & True Positive (TP)            & False Negative (FN)          \\ \cline{2-4} 
    \multicolumn{1}{|c|}{\multirow{-2}{*}{\cellcolor[HTML]{C0C0C0}True Class}} & Negative & False Positive (FP)           & True Negative (TN)           \\ \hline
    \end{tabular}
    \label{tab:confusion-matrix}
    % \vspace{-8pt}
\end{table}

Despite the widespread use of \textbf{accuracy} to evaluate the performance of an algorithm, it may lead to misleading results in imbalanced scenarios and when incorrect classifications have a different cost. 
It is given by the ratio between correctly classified instances and the total number of instances.

% \begin{equation}
%    accuracy = \frac{TP + TN}{TP + FN + FP + TN}
% \end{equation}

\textit{Precision} is given by the fraction of instances classified as positive that are correctly classified: $TP / (TP + FP)$.

\textit{Recall}, also known as \textit{true positive rate (TPR)} or \textit{sensitivity}, is given by the fraction of positive instances that are correctly classified: $TP / (TP + FN)$.

The \textbf{F1-score} corresponds to the harmonic mean of precision and recall, also given by:
\begin{equation}
    F\textit{1}\text{-}score = \frac{2 \times TP}{2 \times TP + FN + FP}
\end{equation}

\textit{Specificity} is usually used alongside sensitivity and is given by the fraction of negative instances that are correctly classified: $TN / (TN + FP)$. \textit{False positive rate (FPR)} is given by $1 - specificity$.

The \textit{receiver operating characteristic (ROC) curve} depicts the trade-off between costs and benefits by plotting the recall against the FPR, as some threshold parameter of the classifier is varied. The \textit{area under the curve (AUC)}, a single quantitative summary of a model's performance~\cite{ESL:HastieTF09}, can be computed from the ROC curve.

\subsection{Fairness Concepts}
\label{ssec:fairness-concepts}

Although the fairness of a software system is difficult to define due to its ambiguity~\cite{Binns18}, throughout this work we consider it to be the absence of bias or discrimination against people based on \textbf{protected or sensitive attributes}, such as race, gender, or age.
% Discrimination based on such characteristics of an individual is usually prohibited by law.
% Unfairness in software systems based on ML may appear in many different forms and may have a variety of root causes. We are particularly interested in the unfairness that might be present in the predictions made by some ML model and its relation with the data used to train the model, which in turn might be biased, as is the case when a dataset is derived from historical discriminatory decisions~\cite{DBLP:conf/www/ZafarVGG17}.
We are particularly interested in the unfairness of the predictions made by ML models, and its relation to the data used to train them, which in turn can be biased if derived from discriminatory historical decisions~\cite{DBLP:conf/www/ZafarVGG17}.

% Models will have \textbf{disparate treatment}, a direct form of discrimination, if there is a deliberated use of the sensitive attribute. In such a scenario, changing the value of the sensitive attribute would result in a different prediction for an individual~\cite{DBLP:conf/www/ZafarVGG17}. This type of discrimination can be avoided by removing the sensitive attribute from the data prior to training the model~\cite{Xu2018}. Even when trained without the sensitive attribute, the predictions might still be discriminatory, leading to an unfair treatment of protected groups~\cite{Calders2010, Xu2018}. This \textit{red-lining effect} is due to the presence of features highly associated with the sensitive attribute~\cite{Calders2010, Xu2018} and is linked to \textbf{disparate impact}. This indirect form of discrimination is not illegal in itself, as long as objective and reasonable justifications to proof the absence of discrimination are given~\cite{RomeiR14,FeldmanFMSV15}. More recently, a novel notion of unfairness, \textbf{disparate mistreatment}, has been proposed by~\cite{DBLP:conf/www/ZafarVGG17}. The rationale behind this notion addresses differences in the misclassification rates of protected and unprotected groups of individuals~\cite{DBLP:conf/www/ZafarVGG17}.
\textbf{Disparate treatment}, a direct form of discrimination, results from a deliberated use of the sensitive attribute and can be avoided by removing it from the data prior to training the model~\cite{Xu2018}. Even when trained without the sensitive attribute, the predictions may still be discriminatory, leading to an unfair treatment of protected groups~\cite{Calders2010, Xu2018}. This \textit{red-lining effect} is due to the presence of features highly associated with the sensitive attribute~\cite{Calders2010, Xu2018} and is linked to \textbf{disparate impact}. This indirect form of discrimination is not illegal in itself, as long as objective and reasonable justifications for it can be given~\cite{RomeiR14,FeldmanFMSV15}. The rationale behind \textbf{disparate mistreatment}, proposed by~\cite{DBLP:conf/www/ZafarVGG17}, addresses differences in the misclassification rates between the protected and the unprotected groups~\cite{DBLP:conf/www/ZafarVGG17}.

In a supervised classification problem, we are given a labelled dataset $\mathcal{D} = \{X, S, Y\}$ of $n$ instances: $X$ are the non-sensitive attributes, $S$ denotes a sensitive attribute, and $Y$ represents the true labels. The variable that represents the classifier's predictions is referred to as $\hat{Y}$.
% A binary sensitive attribute partitions the dataset into two disjoint subsets: the subset composed by the instances for which the value of the sensitive attribute is 0 is called the protected or unprivileged group, while the subset of the instances for which the value of the sensitive attribute is 1 is called the unprotected or privileged group.
A binary $S$ partitions the dataset into the protected or unprivileged group (value of 0 for the sensitive attribute) and the unprotected or privileged group (value of 1 for the sensitive attribute).

% Many \textbf{fairness metrics} have been proposed in the literature, but not all can be applied to the data for which the predicted label is known. We focus on fairness metrics which can be applied at both stages of the pipeline: prior to training a ML model and afterwards, when the predictions are known. The definitions that follow can be applied to the datasets if instead of $\hat{Y}$, the variable representing the predictions, we use $Y$, the variable which represents the true labels.
We focus on fairness metrics which can be applied on different stages: prior to training a model, and afterwards, when the predictions are known. The following definitions can be applied to the datasets, if we use $Y$ instead of $\hat{Y}$.

\textbf{Statistical parity difference}, or the Calders-Verwer score (CVS), considers the difference of the rate of favourable predictions between protected and unprotected groups~\cite{Xu2018}:
\begin{equation}
    P(\hat{Y}=1|S=1) - P(\hat{Y}=1|S=0)
\end{equation}
% Measures of this metric lie in the range $[-1,1]$, with 0 corresponding to optimal fairness. The sign of the measure gives an indication of the skew in favour of the protected or the unprotected group~\cite{Friedler2019}.
Measures of this metric lie in $[-1,1]$, with 0 being optimal fairness. The sign of a measure indicates a skew in favour of either the protected or unprotected group~\cite{Friedler2019}.

\textbf{Disparate impact} (DI) is given by the ratio of the rate of favourable predictions for the protected group to that of the unprotected group~\cite{Friedler2019}:
\begin{equation}
    \frac{P(\hat{Y}=1|S=0)}{P(\hat{Y}=1|S=1)}
\end{equation}
This is often referred to as the $p$\%-rule and for a classifier to be fair, i.e. not to have DI, it should be greater than 80\% but lower than 125\%~\cite{FeldmanFMSV15,ZafarVGG17}. The 80\% rule is advocated by the US Equal Employment Opportunity Commission~\cite{FeldmanFMSV15}, and can be found in the Code of Federal Regulations, in the scope of labour regulations~\cite{cfr}. Measures of this metric lie in $[0,\infty[$, with a value different from 1, the optimal fairness, indicating a skew in favour of one of the groups.

The prejudice index (PI) corresponds to the mutual information between the predictions and the sensitive attribute~\cite{Kamishima2012}. The \textbf{normalised prejudice index} (NPI) results from the application of a normalisation technique for mutual information:
\begin{equation}
    NPI = \frac{PI}{\sqrt{H(\hat{Y})H(S)}}
\end{equation}
where $H(\cdot)$ is an information entropy function~\cite{Kamishima2012}. The NPI ranges between $[0,1]$, with 0 being the optimal value.

\subsection{Related Work}
\label{ssec:related-work}

The approaches that have been proposed to enhance the fairness of ML models or mitigate the bias in their predictions can be grouped into three categories: pre-process, in-process and post-process. \textbf{Pre-process approaches} modify the training data to make it free of discrimination; \textbf{in-process approaches} change the models by adding constraints and regularisation terms to the objective functions; and \textbf{post-process approaches} directly change the predictions made by the models~\cite{Xu2018}.

% Pre-process approaches align with our work in that they explore ways to manipulate the data so that the learning models can be trained with discrimination-free data.
Pre-process approaches align with our work in that they explore ways to manipulate the data before it is used to train the models. The Uniform Sampling and the Preferential Sampling methods proposed in~\cite{Kamiran2012} are similar to those used in our work. In addition to these sampling methods, the authors also propose suppression (removal of the sensitive attribute and those highly correlated with it), massaging (modification of the labelling of the training examples) and reweighing (assignment of weights to the training instances).

We also take inspiration from~\cite{Kamishima2012}, where the authors compare models trained with and without the sensitive attribute, and propose the NPI as a fairness metric. They also consider different data preparation procedures, but no evaluation is performed with regards to using different versions of the same dataset to train the same learning algorithms.

The in-process approaches proposed by~\cite{Kamiran2010} and~\cite{Raff2018} focus on modifying tree-based methods so as to make them discrimination-aware, thus improving the fairness of the models' predictions. This goal is accomplished by changing the evaluation of the splitting criterion and relabelling the leafs.

Some previous work, such as ~\cite{Hardt2016}, \cite{DBLP:conf/www/ZafarVGG17} and \cite{Speicher2018}, focused on the definition of new fairness notions and metrics capable of overcoming some of the known flaws of more traditional metrics, like statistical parity and the 80\% rule.

More recently,~\cite{Friedler2019} focused on defining a benchmark approach to evaluate fairness. A variety of fairness-enhancing methods are compared, and the relation between different fairness metrics is investigated. In contrast, we focus on assessing the impact of standard ML data preparation procedures rather than on fairness-aware methods. This work also alerts to the need to carefully specify the data pre-processing techniques applied to the training data, as they may have a significant impact on the fairness evaluation of a system.

\section{Methodology}
\label{sec:methodology}

The main goal of this work is to \textbf{understand how the different procedures applied to a dataset during data preparation impact the fairness} of the predictions made by a model. 
To achieve this goal we aim at answering the following research questions:

\begin{description}
    \item [RQ1.] How does the removal of the sensitive attribute impact the fairness of the predictions made by an ML model?
    
    \item [RQ2.] How does feature discretisation and the encoding of the categorical attributes impact the fairness of the predictions made by an ML model?
    
    \item [RQ3.] What is the impact of different instance selection techniques on the fairness of the predictions made by an ML model? We consider cross-validators and sampling methods as instance selection techniques.
\end{description}

% We also investigated the impact of these procedures in the models' predictive performance, namely by analysing the trade-off between overall performance and fairness. The impact of the learning algorithm on these two properties of a system was also analysed.
We also investigated the impact of the learning algorithm on the fairness of a system. The predictive performance of the models was also evaluated.

% Our goal is to understand how the different procedures applied to a dataset during pre-processing impact the fairness of the predictions made by a model. This analysis also took into account the effects of these procedures on the model's predictive performance. We considered three types of data pre-processing procedures: discretisation of numerical attributes, methods to deal with imbalanced datasets, and removal of the sensitive attribute.

\begin{figure*}[t!]
	\centering
    \includegraphics[width=.95\linewidth]{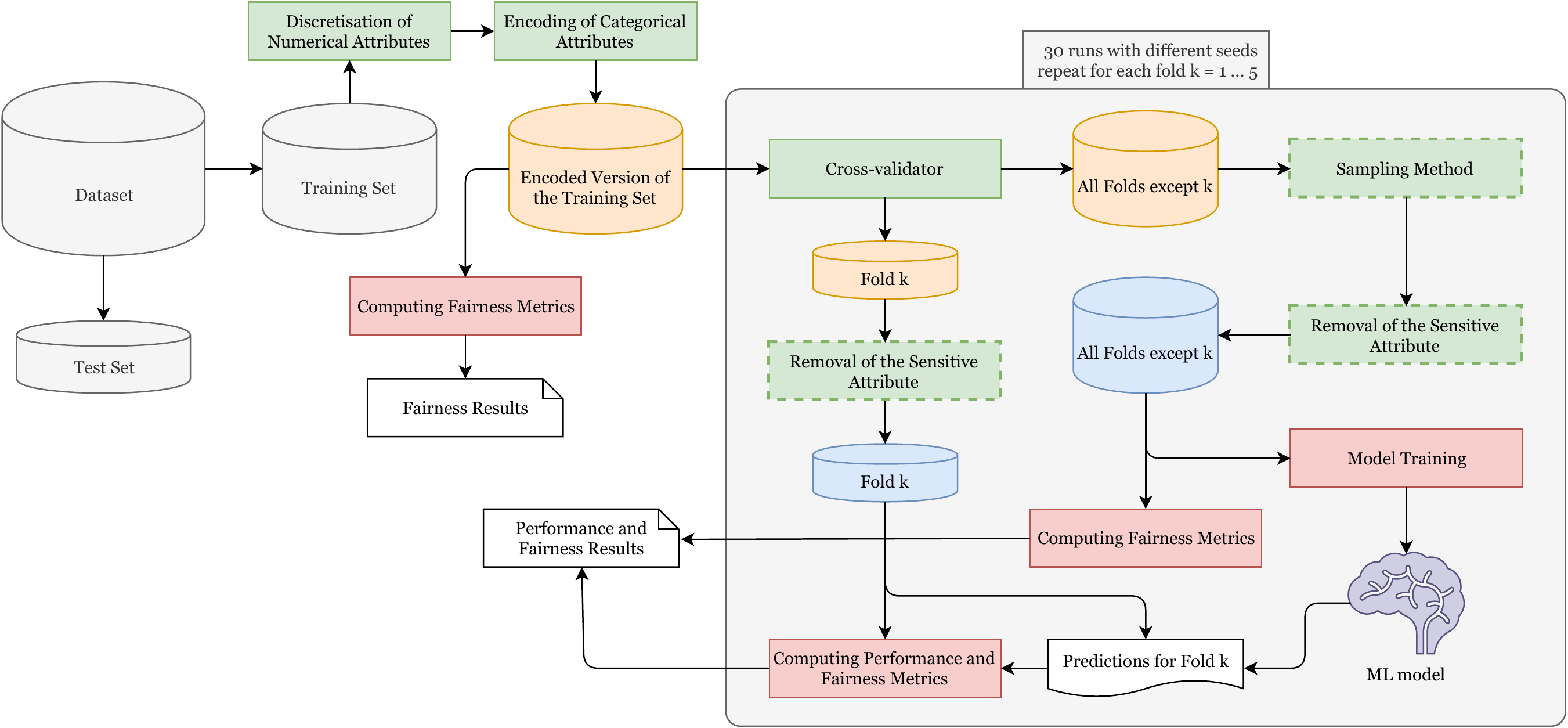}
	\vspace{-8pt}
    \caption{Approach to assess the impact of data preparation on fairness.}
	\label{fig:methodology}
	\vspace{-16pt}
\end{figure*}

Fig.~\ref{fig:methodology} shows the followed approach to assess the impact of data preparation procedures on the fairness of software systems, more precisely on the fairness of the predictions made by an ML model. The steps represented by green boxes are the main focus of this work, with dashed boxes representing optional steps: for instance, under some of the tested configurations, the sensitive attribute is not removed prior to training. These steps are detailed in the remainder of this section. To get a better understanding of the impact of data preparation, we measure fairness at both the data level and the predictions. As depicted in Fig.~\ref{fig:methodology}, we only used the training set of both datasets in our experiments.

% The green boxes are the main focus of this work, even though an analysis is also made with respect to the learning algorithm. The dashed boxes represent optional steps: for instance, under some of the tested configurations, the sensitive attribute is not removed prior to training. These steps are detailed in the remainder of this section. To get a better understanding of the impact of these data preparation steps, we not only measure fairness at the level of the predictions, but also at a data level, after the data suffers some modification. It is important to mention that the experiments only used the training set of both datasets, as depicted in Fig.~\ref{fig:methodology}.

\subsection{Datasets}
\label{ssec:datasets}

We conducted our experiments with two datasets publicly available from the UCI Machine Learning Repository~\cite{UCI-ML}.

The \textit{Adult} or \textit{Census Income dataset} contains demographic data extracted from the 1994 US Census Bureau database, with each instance being described by 14 categorical and numerical attributes. There are 48,842 instances in the dataset and a split into training (32,561 instances) and test (16,281 instances) sets is provided. The main task is to predict whether a person earns over 50,000 dollars per year, therefore making a classification into high or low income. In our experiments, we followed previous work and used \texttt{sex} as the sensitive attribute with \texttt{female} being the unprivileged group.

The \textit{German Credit Data dataset} contains financial information about 1,000 individuals, described by a set of 20 categorical and numerical attributes. The objective is to classify each person into good or bad credit risk. Similar to other studies, we considered \texttt{age} to be the sensitive attribute with \texttt{young} as the unprivileged group, based on the findings reported in~\cite{Kamiran}. The attribute \texttt{sex} can be derived from \texttt{personal-status-sex} of the original dataset. A pre-split into training and test data is not provided for this dataset. Thus, we performed a 70/30 stratified split and tried to maintain the distributions of the true labels and the sensitive attribute on each set.

\subsection{Numerical / Categorical Attributes and Missing Values}
\label{ssec:discretisation-encoding-missing-values}

Following the approaches of~\cite{Calders2010} and~\cite{Kamishima2012}, the numerical attributes were discretised into 4 bins with the boundaries  corresponding to those of the interquartile ranges. An additional transformation was performed for the Adult Income dataset, with bins which correspond to low frequency counts (less than 50 instances) being pooled together and the attribute values being replaced by the same \texttt{Pool} value. This additional transformation was only applied to originally categorical attributes. We refer to this as the integer encoded version of a dataset.

Furthermore, another version of each dataset was created with all features using a one-hot (or 1-of-$K$) encoding scheme~\cite{Bishop2006} after being discretised, meaning that they are represented by binary dummy variables. We refer to this as the one-hot encoded version of a dataset.

Two exceptions to this approach occurred with German Credit Data. The \texttt{personal-status-sex} attribute was removed from both versions of the dataset after deriving \texttt{sex}. The \texttt{age} attribute was discretised into two bins defined by a value greater than or equal to 25, a threshold that was set based on the findings reported by~\cite{Kamiran}.

Contrary to the German Credit Data dataset, Adult Income contains missing values. For the integer encoded version of this dataset, all instances containing at least one missing value were dropped prior to training or testing the models. As far as the one-hot encoded version of the dataset is concerned, all instances were kept regardless of the presence of missing values. In such cases, a missing value was represented by setting all the corresponding dummy variables to zero.

An overview of the Adult Income dataset is shown in Table~\ref{tab:adultd-dataset-analysis}, where the data for the integer encoded version is shown between parenthesis.
For the one-hot encoded version, the unprivileged group only represents around 33.08\% of the dataset. Furthermore, only around 15.04\% of the favourable classifications (\texttt{high-income}) correspond to \texttt{females}. For this version of Adult Income, the CVS is $0.1963$, the NPI is \num{4.35e-02}, and the DI is 35.80\%.

In the integer encoded version of the dataset, \texttt{females} represent around 32.43\% of the training data and around 14.81\% of the favourable classifications, after removing the missing values. The CVS increases to $0.2002$, the NPI suffers a slight increase to \num{4.36e-02}, and the DI is now 36.22\%.
None of the versions of the dataset can be considered fair under the 80\% rule.
The favourable classifications represent 24.08\% and 24.89\% of the training data, for the one-hot and integer encoded versions of Adult Income, respectively.

\begin{table}[htbp]
    \caption{Overview of the one-hot (integer) encoded version of the Adult Income dataset.}
    \centering
    \renewcommand{\arraystretch}{1.15}
    \begin{tabular}{|c|c|C{1.85cm}|C{1.85cm}|}
    \hhline{~|~|-|-|}
    \multicolumn{2}{c|}{}                                                                 & \multicolumn{2}{c|}{\cellcolor[HTML]{C0C0C0}Sensitive Attribute} \\ \cline{3-4} 
    \multicolumn{2}{c|}{\multirow{-2}{*}{}}                                               & Male                      & Female                     \\ \hline
    \multicolumn{1}{|c|}{\cellcolor[HTML]{C0C0C0}}                                        & High income & 6,662 (6,396)        & 1,179 (1,112)          \\ \cline{2-4} 
    \multicolumn{1}{|c|}{\multirow{-2}{*}{\cellcolor[HTML]{C0C0C0}True Label}}         & Low income  & 15,128 (13,984)       & 9,592 (8,670)           \\ \hline
    \end{tabular}
    \label{tab:adultd-dataset-analysis}
    % \vspace{-10pt}
\end{table}

Table~\ref{tab:german-credit-dataset-analysis} presents an overview of the other dataset. In this case, \texttt{young} individuals are represented by 15.00\% of the dataset. The favourable classifications (\texttt{good credit}) represent 70.00\% of the training data, with only 12.65\% of them being assigned to the unprivileged group. For both versions of the dataset, the CVS is $0.1289$, the NPI is \num{9.47e-03}, and the DI is 82.09\% when taking \texttt{age} as the sensitive attribute. According to the 80\% rule, the training set of German Credit Data can be considered fair.

\begin{table}[htbp]
    \caption{Overview of the German Credit Data dataset.}
    \centering
    \renewcommand{\arraystretch}{1.25}
    \begin{tabular}{|c|c|C{1.5cm}|C{1.5cm}|}
    \hhline{~|~|-|-|}
    \multicolumn{2}{c|}{}                                                                 & \multicolumn{2}{c|}{\cellcolor[HTML]{C0C0C0}Sensitive Attribute} \\ \cline{3-4} 
    \multicolumn{2}{c|}{\multirow{-2}{*}{}}                                               & Aged                      & Young                     \\ \hline
    \multicolumn{1}{|c|}{\cellcolor[HTML]{C0C0C0}}                                        & Good credit & 428         & 62          \\ \cline{2-4} 
    \multicolumn{1}{|c|}{\multirow{-2}{*}{\cellcolor[HTML]{C0C0C0}True Label}}         & Bad credit  & 167         & 43           \\ \hline
    \end{tabular}
    \label{tab:german-credit-dataset-analysis}
    \vspace{-10pt}
\end{table}

\subsection{Imbalanced Data and Sampling Methods}
\label{ssec:imbalanced}

The datasets have an imbalance with respect to not only the true labels, but also the sensitive attribute. In such scenarios, it is common to apply sampling methods, namely random undersampling, so as to train the models with an equal number of instances from each class. Besides the typical scenario in which random undersampling is performed w.r.t. the true labels (\texttt{undersampling-label}), we considered two additional configurations: in one it is applied w.r.t. the sensitive attribute (\texttt{undersampling-protected}), and in another it is applied w.r.t. a variable which combines the true labels and the sensitive attribute (\texttt{undersampling-multivariate}).

For each of these settings, we determine which group has fewer instances and keep them, while randomly removing instances from the remaining classes, until their number equals that minimum. After the application of \texttt{undersampling-multivariate} the training data can be considered perfectly fair under the CVS, the NPI, and DI.

We compared these sampling strategies to a baseline scenario in which no sampling method is applied (\texttt{without-resampling}).

\subsection{Learning Algorithms}
\label{ssec:algorithms}

We performed our experiments with Decision Trees and Random Forests. Bearing in mind that we were dealing with categorical attributes, we looked for implementations of these methods which offered support for such attributes. We opted for the implementations provided by the Python API of Apache Spark\textsuperscript{\texttrademark}. We set the maximum depth of the trees to 30 (maximum supported by these implementations) and used the Gini index as the impurity criterion. Additionally, for the Random Forests, we considered ensembles of 10 trees and the squared root of the total number of features when looking for the best split. Our goal was not to fine tune the parameters, but to understand how the different combinations of data preparation techniques and classifiers impacted the system from a fairness point-of-view. Therefore, we used the default values for most of the remaining parameters.

Since the objective is also to analyse the impact of the removal of the sensitive attribute prior to training a model, we devised four possible scenarios: Decision Tree with and without the sensitive attribute (DT and DTns, respectively), and Random Forest with and without this attribute (RF and RFns, respectively).

\subsection{Model Assessment}
\label{ssec:model-assessment}

We performed five-fold cross-validation with the help of the methods provided by Scikit-learn~\cite{scikit-learn}. In addition to the standard version of cross-validation (\texttt{normal-cv}), the experiments were repeated with stratification (\texttt{stratified-cv}) so as to maintain the class distributions of the original data. Furthermore, each configuration was run with 30 different seeds for the random generators.

% We selected fairness metrics which can be applied to the datasets and to the predictions made by the models. By selecting such metrics, we were able to compare the unfairness in the predictions to that originally found in the training data. The selected set of metrics, which definitions can be found in \ref{sec:background-related-work}, includes statistical parity difference (CVS), disparate impact (DI), and normalised prejudice index (NPI).
We selected fairness metrics which can be applied to the datasets and to the predictions made by the models, so as to be able to compare the unfairness in the predictions to that originally found in the training data. The selected set of metrics includes statistical parity difference (CVS), disparate impact (DI), and the normalised prejudice index (NPI).

The F1-score is more suitable when dealing with imbalanced datasets. However, we also include accuracy in our analysis to facilitate the comparison with previous work.

\begin{figure*}[htb]
	\begin{center}
    \includegraphics[width=\linewidth]{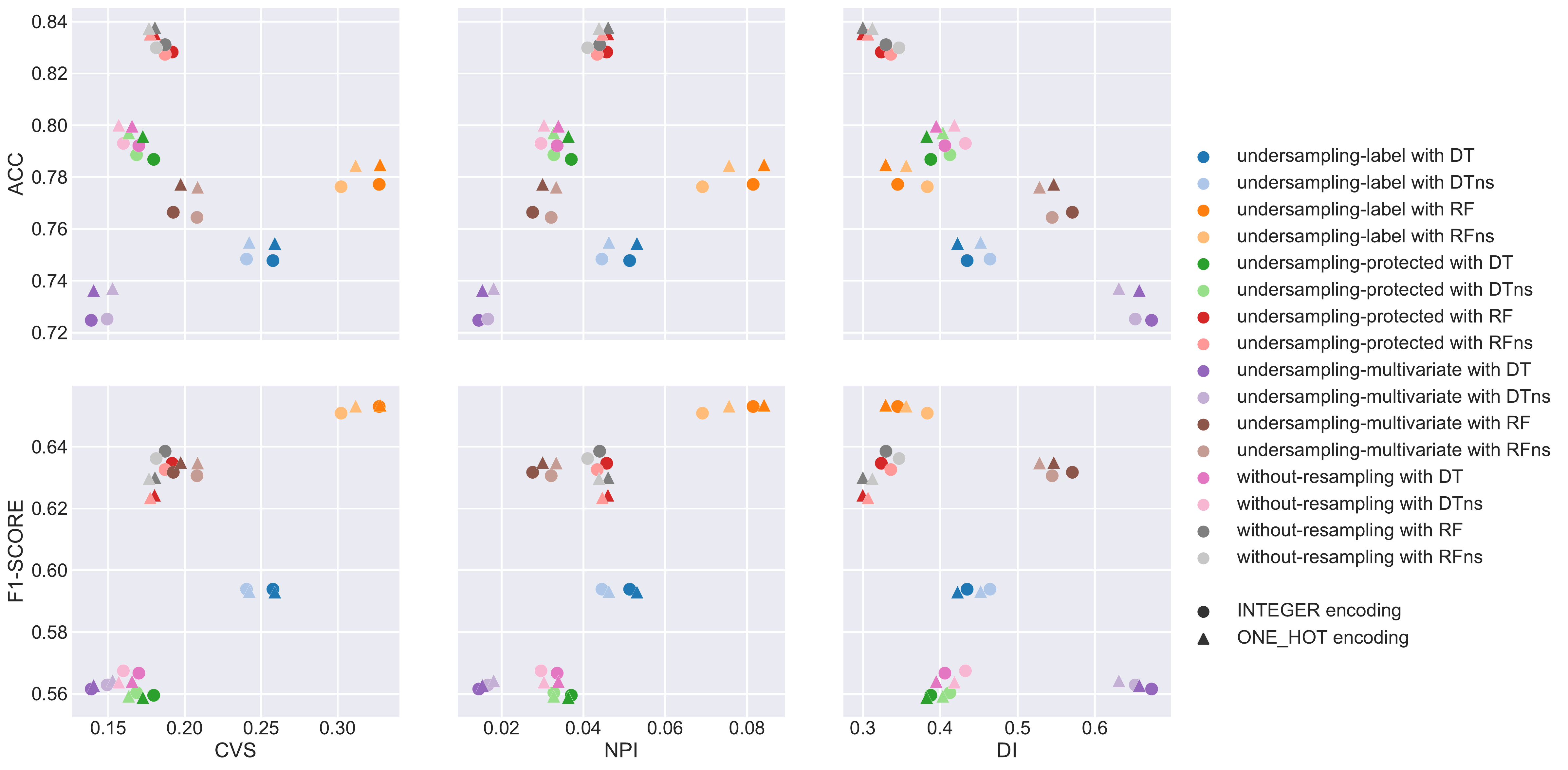}
    \vspace{-16pt}
    \caption{Average predictive performance and fairness when using stratified folds with the Adult Income dataset. From left to right, fairness is given by CVS, the NPI, and DI. Predictive performance is given by accuracy (ACC) in the top row and by the F1-score in the bottom row.}
	\label{fig:adult-income-stratified-cv-scatter-plots}
	\end{center}
	\vspace{-16pt}
\end{figure*}

\section{Results}
\label{sec:results}
We first make a relative comparison between all the tested configurations, both in terms of fairness and predictive performance, and then analyse how the fairness in the predictions relates to the fairness in the training data.

\subsection{Analysis of Fairness and Performance}
\label{ssec:results-fairness-performance}
We remind you that for CVS and the NPI the fairer results are closer to zero, while for DI they are closer to one.

% Fig.~\ref{fig:adult-income-stratified-cv-scatter-plots} shows the average predictive performance and fairness, according to the two performance metrics and the three fairness metrics that were selected, for the Adult Income dataset.
Fig.~\ref{fig:adult-income-stratified-cv-scatter-plots} shows the average predictive performance and fairness for Adult Income, when training the models with a \texttt{stratified-cv}. 
% The results with \texttt{normal-cv} are omitted since they lead to similar conclusions. Thus, we can conclude that, for this dataset, the cross-validator has an almost negligible impact on fairness.
For this dataset, the results with a \texttt{normal-cv} are omitted since the impact on fairness is almost negligible and we reach similar conclusions.

Removing the sensitive attribute prior to training results in models which make fairer predictions under all fairness metrics.
Surprisingly, an exception to this behaviour is observed when applying \texttt{undersampling-multivariate} with both Decision Trees and Random Forests. 
% Another aspect worth mentioning is that if we perform our analysis based on DI, no configuration seems to lead to a fair model, according to the 80\% rule.
It is also worth mentioning that none of the models can be considered fair according to the 80\% rule.

The best sampling method depends on the fairness metric.
% Using NPI, \texttt{undersampling-label} seems to be the worst choice, while \texttt{undersampling-multivariate} seems to be the best option.
Using the NPI, \texttt{undersampling-label} and \texttt{undersampling-multivariate} seem to be the worst and the best option, respectively. Performing \texttt{undersampling-protected} seems to create models whose predictions are more unfair than \texttt{without-resampling}.
Using DI, the sampling methods, from best to worst, are: \texttt{undersampling-multivariate}, \texttt{undersampling-label}, \texttt{without-resampling}, and \texttt{undersampling-protected}.
Using CVS, the sampling methods to train Decision Trees, from best to worst, are: \texttt{undersampling-multivariate}, \texttt{without-resampling}, \texttt{undersampling-protected}, and \texttt{undersampling-label}. On the other hand, to train Random Forests, following the same order, we have: \texttt{without-resampling}, \texttt{undersampling-protected}, \texttt{undersampling-multivariate}, and \texttt{undersampling-label}.

The best encoding depends on the fairness metric and the learning algorithm.
Using DI, the unfairness in the predictions made by models trained with integer encoded data tends to be lower than in the ones made by models trained with one-hot encoded data.
An analysis based on the NPI suggests that all models, except for DT and DTns combined with \texttt{undersampling-protected} or \texttt{without-resampling}, may be able to make fairer predictions if trained with integer encoded. According to CVS, the unfairness in the predictions made by models trained with \texttt{undersampling-protected} and \texttt{without-resampling} in combination with integer encoded data appear to be higher than if those models are trained with one-hot encoded data. The opposite happens with RFns trained with \texttt{undersampling-label}. In fact, most of these differences seem negligible when using the NPI or CVS, except for RFns with \texttt{undersampling-label}.
% The opposite seems to happen when combining DT and DTns with \texttt{undersampling-label} or \texttt{undersampling-multivariate}, RFns with \texttt{undersampling-label}, or RF with \texttt{undersampling-multivariate}, but these differences seem negligible.
% The effect of the encoding seems negligible for RF with \texttt{undersampling-label} and RFns with \texttt{undersampling-multivariate}.

% As far as the learning algorithm is concerned, predictions made by models based on Random Forests are always more unfair than those made by models based on Decision Trees. However, for both \textit{F1-score} and \textit{accuracy}, using Random Forests instead of Decisions Trees seems to be the decisive factor to achieve a better predictive performance.

Measuring performance with F1-score, the results suggest that the second most important factor after the learning algorithm is the sampling method, with models trained with \texttt{undersampling-label} outperforming models trained with any of the other sampling methods that were tested.
The encoding of the categorical attributes also seems to influence the models' performance.
In most cases, a one-hot encoding leads to worse F1-scores than integer encoding.
The exceptions to this behaviour occur when \texttt{undersampling-multivariate} is applied, regardless of the learning algorithm, and when \texttt{undersampling-label} is applied to the data used to train RF and RFns.

\begin{figure*}[t!]
	\begin{center}
    \includegraphics[width=\linewidth]{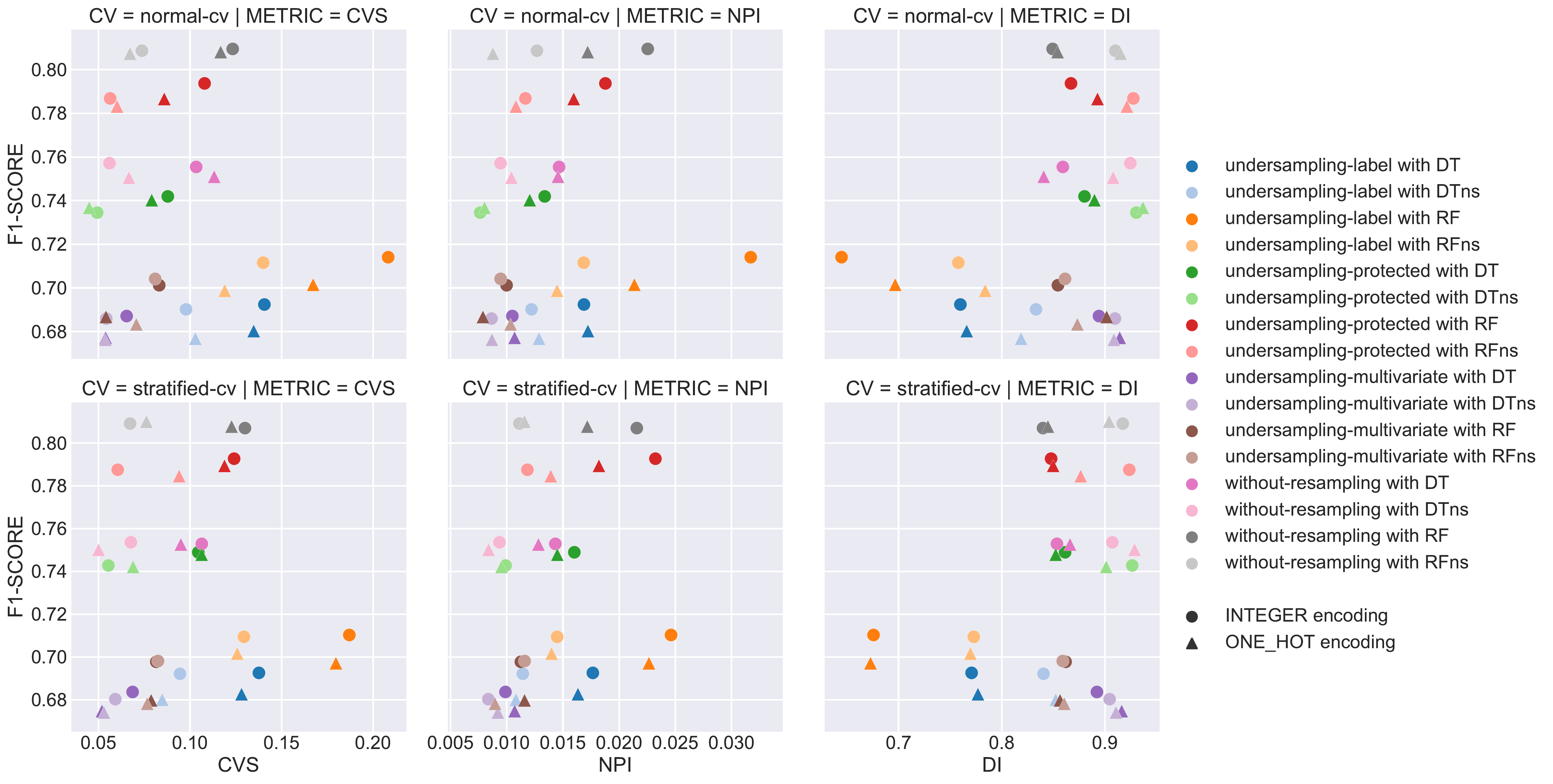}
    \vspace{-16pt}
    \caption{Average F1-score and fairness of all tested configurations with the German Credit Data dataset. From left to right, fairness is given by CVS, the NPI, and DI. In the top row, models were trained with a \texttt{normal-cv} and in the bottom row with a \texttt{stratified-cv}.}
	\label{fig:german-credit-F1-score-scatter-plots}
	\end{center}
	\vspace{-16pt}
\end{figure*}

% Regarding the sampling strategy, \texttt{undersampling-protected} and (\texttt{without-resampling}) have similar effects in terms of accuracy and appear to be the best options. For the other two strategies, performing \texttt{undersampling-label} is likely to be a better choice than performing \texttt{undersampling-multivariate}.
Regarding accuracy, the best options for the sampling method appear to be \texttt{undersampling-protected} or \texttt{without-resampling}, followed by \texttt{undersampling-label}, which seems likely to be a better choice than \texttt{undersampling-multivariate}.
The results also suggest that, regardless of sampling strategy, training models with one-hot encoded data is preferable over an integer encoding.

Fig.~\ref{fig:german-credit-F1-score-scatter-plots} shows the average F1-score and fairness for German Credit Data. The analysis with accuracy leads to similar conclusions regarding the models' predictive performance. For this reason, these results are not shown here.

% Removing the sensitive attribute prior to training and choosing Decision Trees over Random Forests seem to have a more pivotal role on improving fairness than the instance selection methods.
Removing the sensitive attribute prior to training seems to have a more pivotal role on improving fairness than the remaining factors under evaluation. The only exceptions regarding this procedure occur with \texttt{undersampling-multivariate}. The most relevant of which is observed when combining a \texttt{normal-cv} with one-hot encoded data. In this case, the predictions of RFns are more unfair than RF.
% The most relevant of these exceptions happens with Random Forests trained with the one-hot encoded data, with the model trained without the sensitive attribute being more unfair, under all metrics, than the one trained with this attribute.

Moreover, models trained with the one-hot version of the dataset tend to produce fairer predictions than those trained with integer encoded data.
A clear example of this behaviour occurs when combining RF and \texttt{undersampling-label} while using a \texttt{normal-cv}.
Nevertheless, there are exceptions to this behaviour.
When using a \texttt{stratified-cv} and combining RFns with \texttt{undersampling-protected} or \texttt{without-resampling}, the integer encoded version of the dataset actually seems to lead to fairer models.

% In summary, performing \texttt{undersampling-label} in combination with Random Forests seems to be the worst of the tested configurations. Decision Trees produce, in general, fairer predictions than Random Forests. Therefore, model complexity may be a problem for fairness and needs to be further investigated. Additionally, the standard procedure of removing the sensitive attribute prior to training the models seems to be quite effective in improving fairness, while performing \texttt{undersampling-label} might actually lead to a more discriminatory model.
Performing \texttt{undersampling-label} with Random Forests seems to be the worst of the tested configurations, with this sampling strategy often being the worst choice from a fairness point-of-view.

% Decision Trees produce, in general, fairer predictions than Random Forests. Therefore, model complexity may be a problem for fairness and needs to be further investigated.

As far as predictive performance is concerned, models based on Random Forests tend to be better than models based on Decision Trees. Another interesting observation is the effect on predictive performance that the sampling method seems to introduce: applying no sampling method seems to be the best choice, followed by performing \texttt{undersampling-protected}. Performing \texttt{undersampling-label} or \texttt{undersampling-multivariate} leads to worse results, with a less significant improvement between Decision Trees and Random Forests. Nevertheless, moving from Decision Trees to Random Forests has a greater impact on performance than moving from \texttt{undersampling-protected} to \texttt{without-resampling}.

\subsection{Fairness Comparison between Data and Predictions}
% BOXPLOTS ANALYSIS
Besides performing our analysis based on the fairness metrics mentioned in~\ref{ssec:model-assessment}, we computed the ratio between the CVS in the predictions and the CVS found in the data subset used to train the models (CVS Ratio), as well as a similar ratio regarding the NPI (NPI Ratio).
These ratios give an indication of whether the unfairness in the training data was increased or reduced under each configuration.
A value of 1 indicates that the unfairness in the predictions is the same as in the training data, an absolute value greater than 1 means that the unfairness in the predictions is greater, and an absolute value lower than 1 means that the model makes fairer predictions than the procedure which produced the true labels of the training data.
A DI Ratio was not computed since it would be difficult to interpret the results.

\begin{figure*}[t!]
	\begin{center}
        \includegraphics[width=\linewidth]{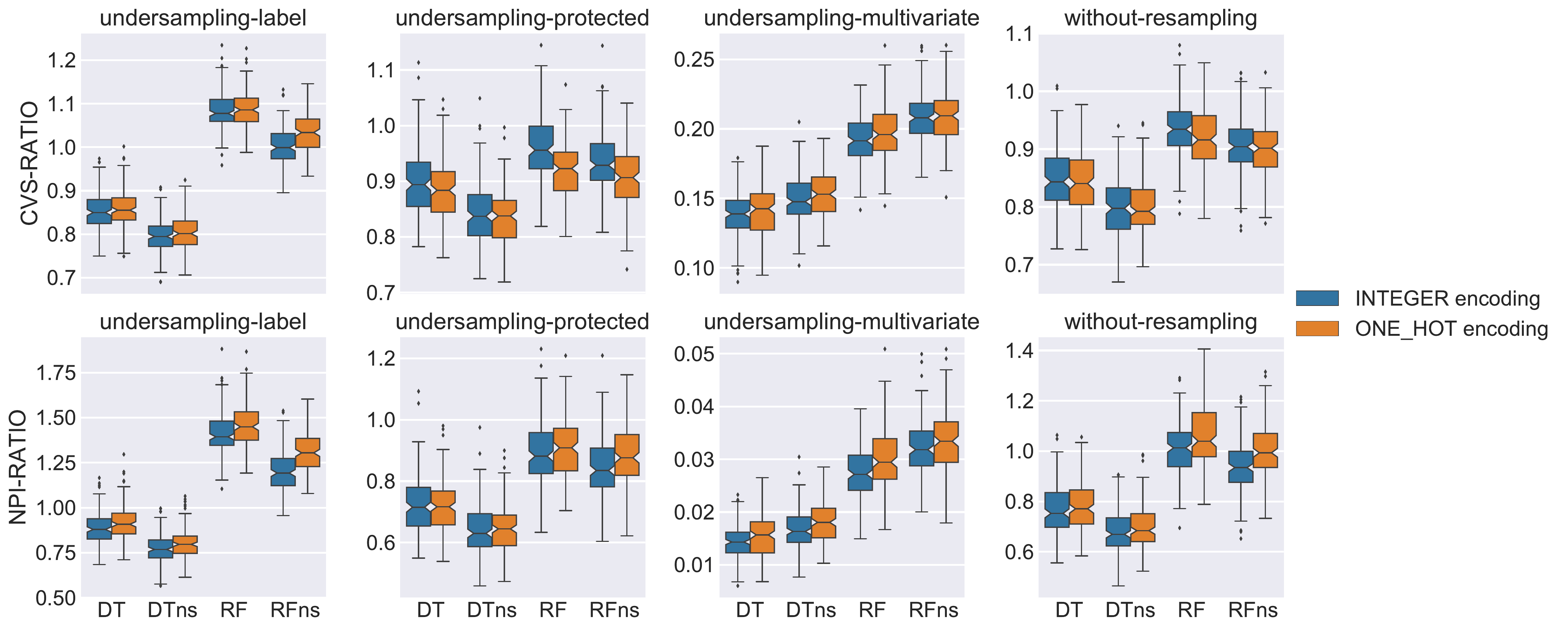}
        \vspace{-16pt}
        \caption{Distributions of the CVS Ratio (top row) and the NPI Ratio (bottom row) for the Adult Income dataset.}
	    \label{fig:adult-income-boxplots}
	\end{center}
	\vspace{-16pt}
\end{figure*}

Caution must be taken when computing these ratios for models resulting from the application of \texttt{undersampling-multivariate}, since the subsets of data used to train these models have a CVS and an NPI equal to zero. In such cases, the value represented in the boxplots corresponds to the CVS or the NPI in the predictions instead of the invalid ratio.

% BOXPLOTS FOR ADULT INCOME
The boxplots in Fig.~\ref{fig:adult-income-boxplots} represent the distributions of the CVS Ratio and the NPI Ratio when a \texttt{stratified-cv} was used with Adult Income. Similar results were observed with a \texttt{normal-cv} and, for that reason, are not presented here.

% BOXPLOTS FOR ADULT INCOME (CVS RATIO)
The CVS Ratio suggests that performing \texttt{undersampling-protected} or not performing random undersampling at all (\texttt{without-resampling}) has similar effects on fairness, allowing for the creation of models that tend to reduce the unfairness in the training data. The opposite happens with models trained with \texttt{undersampling-multivariate} which always increase it. However, the average CVS of the training data with \texttt{undersampling-protected} and \texttt{without-resampling} is around 0.1915 and 0.2050, and so, the unfairness in the predictions is approximately the same between models trained with any of the three sampling strategies.

When applying \texttt{undersampling-label}, caution must be taken when choosing the learning algorithm, since DT and DTns tend to reduce the unfairness of their predictions, while RF and RFns tend to increase it.

% BOXPLOTS FOR ADULT INCOME (NPI RATIO)
The NPI Ratio suggests that the unfairness in the predictions of models trained with \texttt{undersampling-protected} is smaller than the one in the data used to train them. Similarly to what was observed with the CVS Ratio, the unfairness found in the predictions of models trained with \texttt{undersampling-multivariate} is larger than the one in the training data. The unfairness in the predictions of models trained with \texttt{undersampling-protected} and \texttt{undersampling-multivariate} is approximately the same, since the NPI of the training data with \texttt{undersampling-protected} is, on average, between 0.0485 and 0.0534.

When it comes to \texttt{undersampling-label} and \texttt{without-resampling}, DT and DTns tend to reduce the unfairness in the data used to train them. On the other hand, the combination of \texttt{undersampling-label} with RF and RFns tends to result in models whose predictions increase the unfairness in the training data, while the NPI of the predictions made by RF and RFns trained \texttt{without-resampling} tends to be closer to the NPI of the data.

\begin{figure*}[t!]
	\begin{center}
        \includegraphics[width=\linewidth]{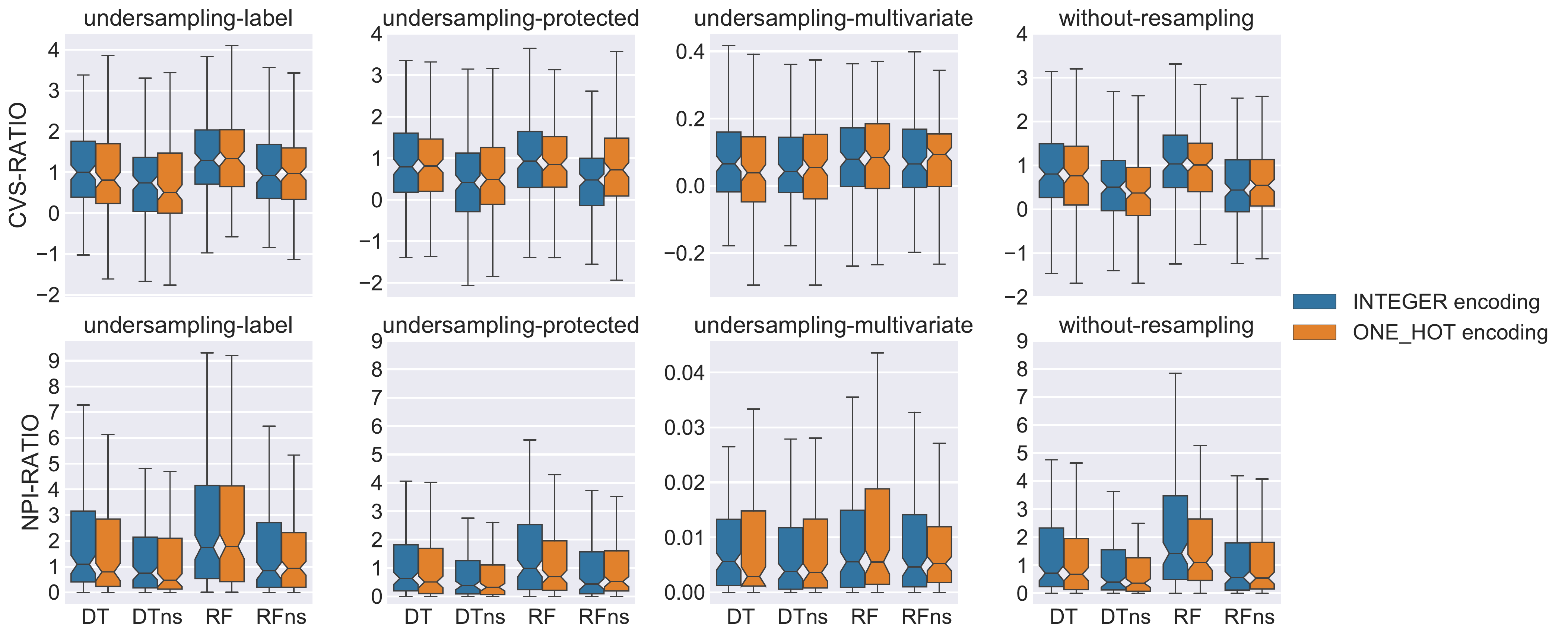}
        \vspace{-16pt}
        \caption{Distributions of the CVS Ratio (top row) and the NPI Ratio (bottom row) for the German Credit Data dataset.}
	    \label{fig:german-credit-boxplots-without-outliers}
	\end{center}
	\vspace{-16pt}
\end{figure*}

% BOXPLOTS FOR GERMAN CREDIT
Fig.~\ref{fig:german-credit-boxplots-without-outliers} represents the distributions of the ratios when a \texttt{stratified-cv} was used with German Credit Data. Extreme outliers, mainly detected with the NPI Ratio, are not represented to allow for a better visualisation. Unless stated otherwise, similar results were obtained with a \texttt{normal-cv}.

% BOXPLOTS FOR GERMAN CREDIT (CVS RATIO)
Similar to Adult Income, applying \texttt{undersampling-multivariate} always results in models whose predictions are more unfair than the data used to train them.

When it comes to \texttt{without-resampling}, the results for the CVS Ratio suggest that the predictions of DT and models trained without the sensitive attribute (DTns and RFns) tend to be more fair than the training data. However, the unfairness of the predictions made by RF may be closer to or higher than that of the training data.

The unfairness in the predictions made by models trained with \texttt{undersampling-protected} tends to be lower than that in the data used to train them. However, in the particular case of stratified folds, the unfairness of the predictions made by models trained with the sensitive attribute may be closer to that of the training data.

With \texttt{undersampling-label}, RF tend to make predictions with an higher unfairness than that of the training data, while those made by DTns tend to have a lower unfairness. This is also observed with the NPI Ratio.
The behaviour of DT and RFns is identical, with their predictions tending to be as unfair as the training data.
The encoding seems to have more impact with a \texttt{normal-cv} than with a \texttt{stratified-cv}: DT and RFns trained with integer encoded data may increase the unfairness in the data, while those trained with one-hot encoded data may reduce it.

% BOXPLOTS FOR GERMAN CREDIT (NPI RATIO)
The results for the NPI Ratio suggest that DT, DTns, and RFns trained with \texttt{undersampling-protected} or \texttt{without-resampling} are able to reduce the unfairness in the training data. However, RF trained \texttt{without-resampling} tend to increase the unfairness in the training data, while RF trained with \texttt{undersampling-protected} tend to reduce it. The only exception, not observed with a \texttt{normal-cv}, are RF trained with integer encoded data for which the unfairness in the predictions is similar to that of the data used to train them.

Regarding \texttt{undersampling-label}, DT trained with integer encoded data tend to make predictions more unfair than the training data, while if trained with one-hot encoding their predictions tend to be fairer. A similar behaviour is observed when combining RFns with a \texttt{normal-cv}. However, when combining RFns with a \texttt{stratified-cv}, the unfairness in the predictions tends to be lower than that in the integer encoded data used to train them, but closer to the unfairness in the one-hot encoded data.

% Even though applying \texttt{undersampling-multivariate} usually creates models for which the unfairness in the predictions is higher in comparison to that in the training data, this unfairness in the models' predictions is likely to be lower than that found in the data used to train models with any of the other three sampling strategies. This is true regardless of the fairness metric used to compute the ratio.
% Although \texttt{undersampling-multivariate} usually creates models for which the unfairness in the predictions is higher than in the training data, the former is likely to be lower than that found in the data used to train models with any of the other sampling strategies. This is true for both ratios.

% CVS RATIO + NPI RATIO
The analysis of the distributions of the CVS Ratio and the NPI Ratio leads to some conslusions that are also supported by the results presented in Section~\ref{ssec:results-fairness-performance}. The unfairness of the predictions made by models trained with \texttt{undersampling-protected} tends to be lower than the unfairness in the data used to train them. These results are similar to those obtained \texttt{without-resampling}, but some configurations may be worse with the latter. Even though \texttt{undersampling-multivariate} increases the unfairness in comparison to the training data, the unfairness in the predictions is similar to that of models trained with \texttt{undersampling-protected}. The worst configuration seems to be the combination of \texttt{undersampling-label} with Random Forests.
% For Adult Income, these ratios also reinforce the unexpected and counter-intuitive behaviour exhibited by models trained with \texttt{undersampling-multivariate} and without the sensitive attribute (DTns and RFns) whose predictions are more unfair than those of DT and RF.
The discrepancies between the CVS Ratio and the NPI Ratio are more accentuated with the smaller dataset.

\section{Discussion and Recommendations}
\label{sec:discussion}

% REMOVAL OF THE SENSITIVE ATTRIBUTE
The \textbf{removal of the sensitive attribute}, together with the learning algorithm, \textbf{is one of the factors that influence the fairness of an ML model the most}. However, removing this attribute so that the learning algorithms do not have direct access to it does not always lead to models that make fairer predictions.
This behaviour, usually exhibited when performing \texttt{undersampling-multivariate}, is somewhat counter-intuitive. When we perform \texttt{undersampling-multivariate}, the training data can be considered fair, at least according to the data level metrics. In addition, we are removing the main source of direct unfairness, i.e. the sensitive attribute. 
We believe this behaviour is caused by indirect prejudice, due to the presence of other attributes highly associated with the sensitive attribute. This indirect prejudice may become more apparent after the removal of the sensitive attribute, an hypothesis that requires further investigation by looking at the structure of the resulting trees.

The removal of the sensitive attribute seems to have low impact on predictive performance, with the models actually having better performance in some cases. Thus, in general, we can consider that this data preparation procedure does not penalize performance to gain in fairness.

% ENCODING
For Adult Income, the choice of the better \textbf{encoding of the categorical attributes} depends on the fairness metric and the learning algorithm, but most of the reported differences seem negligible. According to the NPI and DI, opting for an integer encoding seems to be the safest option for the majority of the configurations that were tested. When using CVS as the fairness metric, the results vary greatly, being difficult to find a more general pattern. Despite the occurrence of some exceptions, an integer encoding of the German Credit Data dataset tends to build models capable of making fairer predictions, independently of the fairness metric. There is no clear answer when it comes to the best encoding in terms of the models' performance.

% STRATIFICATION
When performing cross-validation, opting for using \textbf{stratification seems to have a minor impact on the models' fairness}. While no relevant differences were found for Adult Income, some minor changes were observed for German Credit Data. This may be due to the dataset's size and the under-representation of some classes. Furthermore, no stratification is made with respect to the sensitive attribute. Contrary to the sampling strategy, whose primary goal is to enhance the models' performance, stratification is meant to maintain the distribution of classes between folds and in comparison to the complete dataset, so as to get more accurate estimates.

% SAMPLING STRATEGY
For Adult Income, the better sampling strategy in terms of fairness is dependent on the metric used to perform the analysis. The results with CVS and the NPI suggest that \texttt{undersampling-label} is likely to lead to the worst results, while the results with DI indicate that the worst choice is \texttt{undersampling-protected}. For German Credit Data, all metrics suggest that \texttt{undersampling-label} tend to lead to the worst results. Even though some exceptions may occur, \textbf{\texttt{undersampling-multivariate} should be performed if one's aim is to build fairer models}.
As already mentioned, the training data is completely fair after applying \texttt{undersampling-multivariate}, under the selected metrics. However, the learning algorithm is still able to explore the inherent unfairness of the original dataset, making predictions with some degree of unfairness. This behaviour suggests the existence of indirect prejudice and shows the limitations of the fairness metrics applied at the data level.

The sampling method is actually one of the factors that affects the predictive performance of the models the most. However, for Adult Income, the best sampling method varies with performance metric. Taking the F1-score as a more suitable metric, \texttt{undersampling-label} is definitively the best strategy. As for German Credit Data, performing no random undersampling seems to be the best choice. The behaviour resulting from the application of \texttt{undersampling-multivariate} appears to be quite unstable, but its negative impact may be justified by the more drastic reduction in the number of instances used to train the models.

% LEARNING ALGORITHM
In general, \textbf{models based on Decision Trees produce fairer predictions than those based on Random Forests}, which means that model complexity may be a problem for fairness and needs to be further investigated. A more in-depth analysis of the resulting trees could allow for a better understanding of this behaviour, but we believe it may be due to the randomisation introduced by Random Forests during splitting. However, \textbf{using Random Forests instead of Decision Trees seems to be the decisive factor to achieve a better predictive performance}, which is expected. These observations highlight the trade-offs an organization may face when deploying a model into production.

% GENERAL REMARKS ABOUT FAIRNESS METRICS
An analysis based on the 80\% rule, and the consequent decision on whether to consider a model to be fair, is highly dependent on the dataset. We can illustrate this by comparing the results on the two datasets used in our experiments: for Adult Income, no configuration allows for the creation of a fair model, while for German Credit Data most of the tested configurations originated fair models. We would also like to emphasize that when dealing with data imbalance, it is very unlikely to find a dataset with optimal NPI, since this metric is quite sensitive to small changes in class distributions. The sensitivity of this metric is also exacerbated by the presence of extreme outliers, mainly for the smaller dataset.

% OVERALL RECOMMENDATION
Based on our observations, we would suggest opting for Decision Trees and for following the standard procedure of removing the sensitive attribute to build fairer models. Even though performing random undersampling w.r.t. both the true labels and the sensitive attribute can lead to satisfactory fairness results, it may have a significant impact on the models' performance. \textbf{The best encoding of categorical attributes is data-dependent and different possibilities should be evaluated} instead of choosing an encoding \textit{a priori}. In imbalanced scenarios, stratification is recommended, not only because of it being a good practice, but for its minimal impact on fairness.
Combining \texttt{undersampling-label} with Random Forests should be avoided since other configurations are likely to offer a better trade-off between predictive performance and fairness.

% PERFORMANCE METRICS + IMBALANCED DATASETS
Caution must be taken with the choice of performance metric, especially when dealing with imbalanced datasets, as could be observed with Adult Income. When fairness concerns are also being considered, one should analyse not only the class imbalance with respect to the true labels, as typically done, but also the disproportion between privileged and unprivileged groups. One should also bear in mind that a false negative (for instance, some person being incorrectly classified as bad credit risk) is sometimes more costly than a false positive.

Although being aware of the drawbacks of fairness metrics like statistical parity, as these have been widely discussed in the literature~\cite{Hardt2016,Binns18}, we wanted to perform a not so common analysis of fairness that allowed us to compare the unfairness found in the predictions made by an ML model to that found in the data used to train that model.
Nevertheless, our experiments have shown the brittleness of these metrics, as even those which were expected to show similar behaviours, such as CVS and DI, sometimes presented contradictory results~\cite{Friedler2019}.
This gives further confirmation that there is still room for improvement and progress when it comes to defining new fairness metrics. Specially in imbalanced scenarios, we believe that adopting more recently proposed fairness metrics based on group-conditioned performance~\cite{Friedler2019} might be a small but crucial step towards achieving this goal. There is also room for improvement when it comes to the definition of individual fairness metrics, which seeks to treat similar individuals in a similar way~\cite{DworkHPRZ12}. The challenge here is in finding a suitable measure of the similarity between individuals~\cite{DworkHPRZ12}.

\section{Conclusion and Future Work}
\label{sec:conclusion}

Our goal was to assess the potential impact that data preparation techniques commonly used in a machine learning pipeline may have on the fairness of a software system.
Rather than focusing on fairness-aware methods, we first tried to understand how standard procedures influence fairness, as these are often used to train models without taking fairness concerns into account.
Our findings suggest that the removal of the sensitive attribute and the learning algorithm are the factors that impact the fairness of a system the most.
Despite potentially improving the overall performance of a model in imbalanced contexts, random undersampling must be performed with caution, since it may negatively impact the system's fairness.
As future work, we plan to analyse the structure of the resulting trees, without neglecting the characteristics of the data used to train the models.
We also want to extend the analysis to a wider range of fairness metrics, bearing the possible data imbalance in mind.

\section*{Acknowledgement}
\label{sec:ack}

This work has been partially supported by the project \textbf{ATMOSPHERE} (\href{http://www.atmosphere-eubrazil.eu}{atmosphere-eubrazil.eu}), funded by the Brazilian Ministry of Science, Technology and Innovation (Project 51119 - MCTI/RNP 4th Coordinated Call) and by the European Commission under the Cooperation Programme, Horizon 2020 grant agreement no \texttt{777154}.
It is also partially supported by the project \textbf{METRICS}, funded by the Portuguese Foundation for Science and Techndology (FCT) -- agreement no \texttt{POCI-01-0145-FEDER-032504}. 

\newpage

\bibliographystyle{IEEEtran}
\bibliography{IEEEabrv,references}

\end{document}